\documentclass[runningheads]{llncs}
\usepackage{graphicx}
\usepackage{comment}
\usepackage{url}

\usepackage{times}
\usepackage{url}
\usepackage{amsmath}
\usepackage{color}
\usepackage{todonotes}
\usepackage{multirow}
\usepackage[ruled,linesnumbered]{algorithm2e}
\usepackage{subcaption}

\usepackage{pgfplots}
\usepackage{tikz}
\usepackage{float}

\begin{document}
\title{Knowledge Graph Refinement based on Triplet BERT-Networks\thanks{{\text Copyright~\copyright~2022 for this paper by its authors. Use permitted under Creative Commons License Attribution 4.0 International (CC BY 4.0).}}}
%\titlerunning{Abbreviated paper title}
% If the paper title is too long for the running head, you can set
% an abbreviated paper title here
%
\author{Armita Khajeh Nassiri\inst{1}\orcidID{0000-0002-5734-0351} \and
Nathalie Pernelle\inst{1,2}\orcidID{0000-0003-1487-393X} \and
Fatiha Sa\"{i}s \inst{1}\orcidID{0000-0002-6995-2785}\and Gianluca Quercini\inst{1}\orcidID{0000-0001-9195-1618} }
\authorrunning{Khajeh Nassiri et al.}
% First names are abbreviated in the running head.
% If there are more than two authors, 'et al.' is used.
%
%\institute{Princeton University, Princeton NJ 08544, USA \and
\institute{
$^1$ LISN, CNRS (UMR 9015), Paris Saclay University, France\\
$^2$ LIPN, CNRS (UMR 7030), University Sorbonne Paris Nord, France \\
\email{firstname.lastname@lri.fr}
}

\maketitle              % typeset the header of the contribution
\begin{abstract}

Knowledge graph embedding techniques are widely used for knowledge graph refinement tasks such as graph completion and triple classification.
These techniques aim at embedding the entities and relations of a Knowledge
Graph (KG) in a low dimensional continuous feature space. 
This paper adopts a transformer-based triplet network creating an embedding space that clusters the information about an entity or relation in the KG. It creates textual sequences from facts and fine-tunes a triplet network of pre-trained transformer-based language models. It adheres to an evaluation paradigm that relies on an efficient spatial semantic search technique. We show that this evaluation protocol is more adapted to a few-shot setting for the relation prediction task. 
Our proposed GilBERT method is evaluated on triplet classification and relation prediction tasks on multiple well-known benchmark knowledge graphs such as FB13, WN11, and FB15K. We show that GilBERT achieves better or comparable results to the state-of-the-art performance on these two refinement tasks.

\end{abstract}

\section{Introduction}
\label{sect::introduction}

Large-scale knowledge graphs (KG) have gained popularity in the industry, like those built by Amazon, Google, and Microsoft, and in academia, such as YAGO and Wikidata.
Knowledge graphs can be described as a structured repository of facts presented as triples (head, relation, tail). 
In other words, the facts are relations held between two entities, for instance (Barack Obama, isMarriedTo, Michel Obama). More recently, KGs have become a natural resource for many AI downstream tasks such as question answering, entity disambiguation, and rule mining \cite{Shaoxiong-2021-a-survery}. Although knowledge graphs contain millions of facts, they are considerably incomplete. Three different refinement tasks pertinent to this issue have attracted substantial attention in the literature: Triple classification, Relation Prediction, and Link Prediction. Most of the approaches proposed for tackling these tasks rely on embedding entities and relations of the KG into vectorized representations. 

Commonly, these embedding techniques \cite{Shaoxiong-2021-a-survery} compute a defined scoring function that, given the fact, computes its plausibility. Hence, at the time of evaluation, depending on the nature of the task, the 
existence likelihood of unobserved facts (through substitution) should be calculated using the scoring function and then ranked.
This causes additional overhead, especially for tasks such as link prediction. Others, like KG-BERT\cite{Yao2019-kgbert}, can tackle the prediction task, e.g., relation prediction, as a classification task. Nevertheless, many relations in KGs are long-tail, meaning that very few facts with that relation exist in the KG. In such few-shot learning settings, the classification will be biased towards the data-rich relations, and the approach's robustness can be hampered.

%To the best of our knowledge, prior to our work, KG-BERT has been the only work to adapt the use of pre-trained transformers for the knowledge graph refinement tasks. 
%Given the success of KG-BERT and the popularity and improvements brought by pre-trained language models to NLP, it is natural to build upon BERT new methods than can be adapted to KGs. 

This paper proposes GilBERT, which employs a triplet architecture of a pre-trained language model such as BERT to cluster, in embedding space, all the facts regarding a relation or an entity. It overcomes the earlier shortcomings by fine-tuning BERT with a triplet loss that takes 
textual sequences created from facts as input. This way, depending on the refinement task, we force the information about an entity or a relation to be embedded close to each other, forming clusters. The evaluation protocol uses an efficient spatial semantic search technique, e.i., FAISS \cite{johnson-2017-Faiss} to compute the closeness of the embedding of a test example to the point(s) in the embedding space. This reduces the overhead caused by substituting the candidates, and as shown in the experiments, it is more suited for a few-shot learning setting in the case of relation prediction. 
In addition, unlike translational and factorization-based models that need extra efforts to integrate any additional information such as entity descriptions, types, or paths to them \cite{Xie_2016_DKRL} \cite{wang-2020-dolores}, it is trivial to do so in our proposed model. We also propose strategies that capture some of the semantics of the KG to generate harder negative information for the triplet network. % and reduces the overhead of previous evaluation techniques.

%is more robust than some previous works (e.g., KG-BERT) and has less overhead. 

%rewrite this part. Before say what is particular and new and then give a brief explanation about the approach itself. 

%what is particular is the negative 

%textual gives some flexibility that some approaches don't have such as the abstract

%- We propose strategies to choose harder negative samples that capture some of the semantics of the KG

%In this paper, we propose a new knowledge graph refinement method that represents information about entities and relations as textual sequences. It uses a transformer based triplet architecture to cluster facts that belong to the same entity or relation close to each other in the space. To the best of our knowledge, we are the first to use such triplet architecture and the proposed evaluation method for tackling the knowledge base refinement tasks of relation prediction and triple classification. Unlike previous works, the evaluation is not performed as a classification task, allowing us to be independent from the number of candidates (e.g., number of relations in the KG for the relation prediction task) and hence can scale to KGs with arbitrary number of properties.
%To train the model, we propose strategies to choose hard negative partial facts.

\section{Related Work}
\label{sect::related-work}

% beginning too mcuh intro, has already been said in the introduction 
%directly dive into the works

%Knowledge graphs have drawn considerable research attention mainly due to their increasing applications in information retrieval, question answering, etc. They have also become an indispensable complement to many machine learning techniques. However, due to their ever-evolving and ever-expanding nature, it is evident that the knowledge graphs are incomplete. Thus addressing downstream tasks such as graph completion and relation prediction is quite important. 

%rewrite

%structure it about the evaluation --> and approaches  
% the importance of hard negatives, usually the negatives 
% add some fun example. 
%NLP --> transformers + historical (h+t=r) and the way most rely on it

% say that we improve on KG bert ...

A plethora of research is devoted to finding embeddings that model entities and relations of a KG in a low dimensional continuous vector space and to the proposition of scoring functions capable of capturing different kinds of relations. A recent comprehensive study overviews these techniques \cite{Shaoxiong-2021-a-survery}.
We can roughly categorize these models into three different categories. First, the translational-based models such as the TransE \cite{Bordes2013-transee} and its more advanced extensions like TransH \cite{Zhen2014-transH}, and TransR \cite{Yankai2015-transR}, and more recently RotatE \cite{sun-2018-rotate}. The translational-based models have proved to be successful from the beginning. TransE proposes one of the simplest and most prominent scoring functions that, given a fact, the head $h$ and tail $t$ embeddings are related by a translation vector $r$ such that $h+r \approx t$. Secondly, the tensor factorization-based models with RESCAL \cite{Nickel2011-rescal} being one of the very earliest works adopting this technique. Other extensions DistMult \cite{yang2015-DistMult}, and ComplEx \cite{trouillon-2017-complEx} instead of the tensor product, use dot product and Hermitian dot product, respectively. Thirdly, the neural network-Based Models like ConvE\cite{dettmers-2018-conve} 
adopt CNNs with deeper scoring functions. Also, more recent transformer-based models such as KG-BERT\cite{Yao2019-kgbert} and R-MeN\cite{Nguyen-2020-RMeN} have been proposed in the literature to solve different downstream tasks. To the best of our knowledge, KG-BERT is the only work that fine-tunes a pre-trained language model like BERT.

%Given the success of KG-BERT and the popularity and improvements brought by pre-trained language models to NLP, it is natural to build upon BERT new methods than can be adapted to KGs. 

As pointed out in Section \ref{sect::introduction}, unlike most of the previous works in the literature where the embeddings for entities and relations are learned, we embed the training data into different clusters in our approach for entities or relations, depending on the downstream task. 
Following the indispensable contribution of transformers and pre-trained language models to NLP and the success of KG-BERT, it is natural to build upon BERT new methods that can be adapted to KGs. GilBERT adopts a triplet architecture that fine-tunes BERT (or any similar language model such as RoBERTa).

\section{The proposed approach}
\label{sect::approach}

%begining of approach more like the end of related work. remove it and dive right into the approach itself

%as the introduction: lay out the steps: based on the tasks you want to tackle, you cluster differently, you have to create the data, pass to model, evaluate

%In this section, for simplicity throughout the rest of the paper, we first describe some useful notations. We will then describe the model, the data to be fed to the model, and how harder and more semantically relevant negative examples are chosen. In the end, we briefly explain the evaluation paradigm and will leave its details based on the refinement task to Section \ref{sect::experiments}.

Let $\mathcal{G}$ be a knowledge graph containing a collection of facts represented as triples $\{(h, r, t)\} \subseteq \mathcal{E} \times \mathcal{R} \times \mathcal{E}$, where $\mathcal{E}$ and $\mathcal{R}$ denote the entity and relation sets, respectively. We propose GilBERT, an approach that fine-tunes BERT such that based on the refinement task, it clusters information about an entity (e.g., for triple classification task) or a relation (e.g., for relation prediction task). 
For some downstream tasks, not all triple parts $(h,r,t)$ are known at the time of evaluation; e.g., the relation is unknown for relation prediction. Hence, to create the embedding space, we only use parts of the facts that are available at the time of evaluation so that the test data can be embedded the same way as during training. 
For the simplicity of notation, 
%and $h$ is the head, $r$ is the relation and $t$ is the tail. 
%All triples that are not in the KG are denoted by $\mathcal{G}'$. 
%The reference set $\Delta$ is the set of embeddings of all training inputs obtained after the model is trained. $\Delta_{r_{s}}$ and $\Delta_{e_{s}}$ is the set of embeddings of all those inputs that form the cluster of a specified relation $r_{s}$ or entity $e_{s}$, respectively. 
%We define "partial facts" to be textual sequences representing information about an entity or relation that will be used to form the clusters.
%These partial facts are created differently based on the refinement task.
let a Partial fact $PF$ be a textual sequence created out of a triple $(h,r,t)$ by concatenating the entities or entities and relations.
For instance, given the triple (Barack Obama, isMarriedTo, Michel Obama), $PF_{rt}$ would be: "isMarriedTo Michel Obama", and $PF_{ht}$ would be "Barack Obama Michel Obama". In all cases, and subject to availability, it is possible to replace the entity with its description to inject more background knowledge into the model.

\begin{figure}[H]
    \centering
    \input{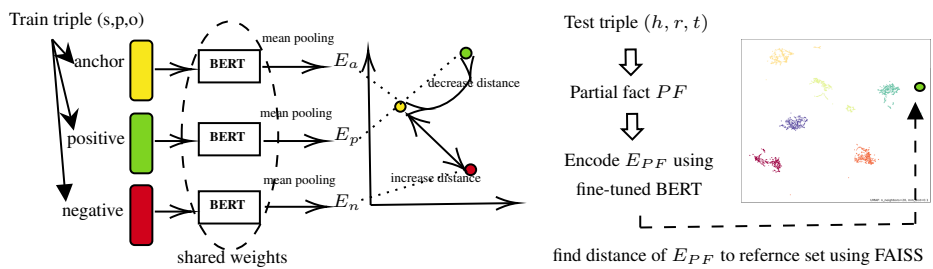}
    \caption{(a) Overview of fine-tuning BERT using the partial facts of triples in $\mathcal{G}$. (b) Overview of evaluation, clusters illustrated using UMAP on part of FB15K dataset for the relation prediction task.}
    \label{Fig:model-eval}
%    ~ 
%    \begin{subfigure}[t]{0.4\textwidth}
%        \centering
%        \input{src/pgfplots__Diagram__2.tex}
%        \caption{Subcaption of the right diagram.}
%        \label{Fig:SubRight}
%    \end{subfigure}
%    \caption{Caption of both diagrams.}
%    \label{Fig:Main}
\end{figure}

%Instead of using the output of the \textsc{[CLS]} token produced in the last layer of these pre-trained transformer based models, a mean pooling layer is applied to their output  which can provide us with a fixed-sized embedding vector given the partial facts as input.

To cluster the training data, inspired by Sentence-BERT \cite{reimers-2019-sentence-bert}, we use a triplet network architecture of BERT that maps semantically similar facts near each other and dissimilar facts distant from one another. Thus it takes as input three partial facts $PF$, one as the anchor, one as positive (which is similar to the anchor), and one as negative (which is dissimilar to the anchor). To obtain a fixed-sized embedding vector for the input $PF$, unlike KG-BERT, which uses the output of the \textsc{[CLS]} token produced in the last layer of BERT, a mean pooling layer is applied to BERT's output layer. According to \cite{reimers-2019-sentence-bert} and our experiments, the output of the pooling layer results in better embedding for the given input than using the embedding of the \textsc{[CLS]} token.
This architecture clusters all the information about an entity (or relation).
%, that is fed to it as partial facts.
%and is more adapted to the few-shot setting. 
It is important to note that using $PF$s as inputs to the model allows us to use only the parts of a fact known at the time of evaluation (e.g., the head and the tail for a relation prediction task). This avoids the evaluation time overhead of previous models caused by creating all possible substitutions and ranking their plausibility based on the learned scoring function. 

Figure \ref{Fig:model-eval} illustrates the fine-tuning of BERT through the triplet network that is comprised of three identical networks with shared parameters. The network is fed three input textual sequences: the partial facts of anchor, positive and negative, with the aim of anchor and positive ending up in the same cluster and anchor and negative in different ones. More specifically, the euclidean distance of the embedding of anchor $E_{a}$ and positive $E_p$ is reduced whilst that of the anchor $E_{a}$ and negative $E_n$ is increased. The triplet loss function is defined as:

$$ \mathcal{L}= max(0, dist(E_{a}, E_{p}) - dist(E_{a}, E_{n}) + \gamma ),$$

%For the triple classification task, the data is created such that the anchor and positive belong to the same entity and the negative is a fact belonging to some other entity. The same holds true for the relation prediction with the difference that the hubs are created for relations. 

\noindent where $\gamma$ is the safety margin ensuring that the embedding of the negative is at least $\gamma$ further from the anchor than the positive.

For each gold triple $(h,r,t)$ in $\mathcal{G}$, $n$ different input anchor, positive and negative $PF$s are created to train the model. It is important to create hard negative inputs so that the function $\mathcal{L}$ is not trivial. In literature, to learn the scoring function that yields embeddings, the negative examples are usually the corruption of a gold triple by replacing either its $h$ or $t$. More sophisticated techniques such as \cite{claudia-2021} utilize background knowledge such as the schema of KG to choose negatives better.
We consider the more frequently observed case where a schema is not available and try to capture the semantics of the KG when generating our input data. Hereunder, we provide the details of data creation depending on whether the clustering is done based on relations or entities.

%The details of creating data and the evaluation differ based on the refinement tasks, as the clusters are created based on entities for one and are based on relation for the other, and are hence detailed separately.
%For the considered gold fact, a number of $n$ different positive and negative samples are created and based on whether the clusters are based on the entity or relation, the data that is passed as input to the model differs. Hence the process is detailed separately.

\noindent\textbf{Clustering based on Relation.}
The anchor is the partial fact $PF_{ht}$. Each positive sample, $PF_{h_{i}t_{i}}$ is created from a fact drawn at random from the set $\{(h_{i}, r, t_{i})\ | (h_{i}, r, t_{i})\in\mathcal{G}\}$, 
and the negative samples $PF_{h_{j}t_{j}}$ from the set $\{(h_{j}, r^\prime, t_{j})\ | (h_{j}, r^\prime, t_{j}) \in \mathcal{G} \wedge r'\in  close_{r} \}$. So for generating hard negatives, the $PF$ is created from a triple whose relation is semantically close to the relation of the anchor and positive. Denote the set $F_{r}$ of all triples in $\mathcal{G}$ having the relation $r$. The set $close_{r}$ is chosen based on two criteria: (1) of all relations in $\mathcal{R}$, the percentage of tail entities shared with $F_{r}$, and (2) the percentage of head entities shared with $F_{r}$. The obtained percentages have the intuition of domain and range, and $close_{r}$ is the set of all relations whose computed sharing percentage is higher than a threshold. For instance, the $close_{BornIn}= \{worksIn, diedIn, locatedIn\}$.

%the sorted of the two sets. A threshold can be set on the values to restrict the choices. % give an example of the closest properties to "bornIn" 

\noindent\textbf{Clustering Based on Entity.}
% head 
The clusters for entities are created using the entities appearing as head, and they can easily be extended to the tail by generating reverse relations. 
%consider that r-1 is added 
The anchor is the partial fact $PF_{rt}$ or randomly $PF_{hrt}$. The positive example is the partial fact created from a triple with the same head as anchor $\{(h, r{i}, t_{i})\ | (h, r_{i}, t_{i}) \in \mathcal{G}\}$. To create negative samples that are more difficult, we adhere to the idea that entities that have the same type are semantically more related to each other. Hence we choose an entity that is likely to have the same type as the head entity $h$. We create $PF_{rt_{j}}$ or $PF_{h_{j}rt_{j}}$ from the set $\{(h{j}, r, t_{j})\ | (h_{j}, r, t_{i}) \in \mathcal{G} \wedge (h, r, t_{j}) \notin \mathcal{G} \}$. This way, rather than setting apart a city from a person, which is too easy for the loss function, we disjoint two people who also share a property.

%More specifically, to perform either of the KG refinement tasks, our approach is comprised of two different steps: (1) data creation and model training, and (2) evaluation. 

%\begin{definition}
%\label{def:partial-fact}
%\textbf{Partial Fact.} The partial fact $PF_{po}$, $PF_{sp}$, or $PF_{sp}$ is the textual sequence that is created given a fact $(s,p,o)$ by combining the subject, property, or object, based on the notation. For instance, to create $PF_{po}$ the textual sequence "p + o" is used.
%together the property and object "p + o", subject and property "s + p", or "s + p", respectively.

%Based on the refinement task, the partial fact of a given triple (s,p,o) is constructed differently. In order to have the embeddings belonging to the same entity close to each other in the embedding space, i.e., to have hubs based on entities, $PF_{e}$ which is the textual sequence "p + o" is used. And to create the hubs based on relations, $PF_{r}$ which is the textual sequence "s + o" is used.

%\end{definition}

%The evaluation is done quite differently and more efficiently than the previous works, where depending on the task, the possible candidate triples are constructed and the dissimilarity score calculated using their scoring function is used to rank the candidates. 

An overview of the evaluation pipeline for the relation prediction task is depicted in Figure \ref{Fig:model-eval}. First, the partial fact of a test triple $(h, r, t)$ is constructed in the same way as it has been done with the gold training inputs. The fine-tuned model is used to obtain the vectorized embedding $E_{test}$. The similarity, through Euclidean distance, of $E_{test}$ to the embeddings of training inputs, is used for the evaluation. Let the reference set $\Delta$ and $\Delta_{e}$ be the set of embeddings of all training anchors and the set of embeddings of anchors corresponding to entity $e$, respectively. 
%The c set $\Delta$ is the set of embeddings of all training inputs obtained after the model is trained. $\Delta_{r_{s}}$ and $\Delta_{e_{s}}$ is the set of embeddings of all those inputs that form the cluster of a specified relation $r_{s}$ or entity $e_{s}$, respectively. 
For the triple classification, the distance of $E_{test}$ is calculated with the reference set of its head entity $\Delta_{h}$. And for relation prediction, the $K$ closest points to $E_{test}$ in $\Delta$ are used for choosing the most plausible relation. In both settings, to find the closest points to $E_{test}$ in the corresponding reference sets, we rely on FAISS \cite{johnson-2017-Faiss}, a library for very efficient similarity search that trades precision for speed.  
More details of evaluation based on each refinement task will be outlined in Section \ref{sect::experiments}. 
% use aggregate based on task to decide finally. 

%The details of evaluation based on each refinement task will be outlined in Section \ref{sect::experiments}. 

\section{Experimental Results}
\label{sect::experiments}

% how long it takes to run/ the evaluation 

\textbf{Datasets}. In this paper, we adopt three widely used benchmark datasets in the literature: WN11 \cite{socher2013reasoning}, FB13 \cite{socher2013reasoning}, and FB15K\cite{Bordes2013-transee}.
%, as well as the recently proposed dataset {C}o{DE}x-M \cite{safavi-koutra-2020-codex}. 
The datasets are subsets of the real-world KGs, Wordnet, and Freebase, respectively. We have used the same training/validation/test splits as \cite{socher2013reasoning}, and \cite{Bordes2013-transee}. 
The test sets of WN11 and FB13 contain both positive and negative triples, whereas the FB15K dataset only contains correct triples.
The statistics of the datasets are listed in Table \ref{tab:stat-data}.

%The negative test triples of the {C}o{DE}x-M dataset are hard negatives whose generation is detailed in \cite{safavi-koutra-2020-codex}. 

%The {C}o{DE}x with only positive triples for validation and test which can be used for the relation prediction task has 10,310 and 10,311 triples respectively. 

\noindent\textbf{Settings.} In our experiments, we have fine-tuned the pre-trained RoBERTa-Base model for 4 epochs. We have used a batch size of 64, Adam optimizer with learning rate 2e-5, and margin $\gamma$ of 5.
It should be noted that we have not used any external information such as paths or entity descriptions while training the model, and doing so can potentially enhance the results. 
All experiments have been run on a single machine with a processor 2.6GHz, 12 cores, 64 GB of RAM, and an Nvidia TITAN V GPU. The code and datasets are publicly available\footnote{{\url{https://github.com/armitakhn/gilbert/}}}.

\begin{table}[!ht]%\setlength\belowcaptionskip{-25pt}
	\centering
\setlength{\tabcolsep}{6pt}

\begin{tabular}{l|lllll}
\hline
Dataset       & \#R  & \#E    & \multicolumn{3}{l}{\#Facts (Train/Valid/Test)} \\ \hline
WN11          & 11   & 38,696 & 112,581         & 2,609         & 10,544        \\
FB13          & 13   & 75,043 & 316,232         & 5,908         & 23,733        \\
FB15K         & 1345 & 14,951 & 483,142         & 50,000        & 59,071        \\
%CoDEx & 51   & 17,050 & 185,584         & 20,620        & 20,622  \\
\hline
\end{tabular}
\caption{Statistics of datasets}
\label{tab:stat-data}
\end{table}

% if we are to keep the codex dataset, add that we consider codex medium with 

\noindent\textbf{Triple Classification.} This task assesses whether a given triplet $(h, r, t)$ is correct or not, i.e., binary classification on a triplet. The evaluation is done on two benchmark datasets of WN11 and FB13. For this task, we cluster based on entities appearing as head in $\mathcal{G}$ as explained in Section \ref{sect::approach}. The partial facts $PF_{hrt}$ are used randomly with a 30\% chance. 
For the WN11 dataset, the reverse versions of relations $(t, r^{-1}, h)$, are added to the KG to possibly overcome the cold start problem caused by some of the test data's heads not appearing as the head to any fact in training. We create five different positive and five negative samples for each anchor in all datasets. %this can be set as a hyperparameter that we say we tuned.

For the evaluation of a triple $(h, r, t)$, the embedding of $PF_{hrt}$'s distance is calculated to 
%all the training points' embeddings forming the cluster for the entity $h$, i.e., 
the reference set $\Delta_{h}$.
%(i.e., to the embedding of all facts in the cluster of the head entity $h$). 
We have exploited different decision strategies, \textsc{Mean}, \textsc{Max}, and \textsc{Min}, to aggregate the computed distances to the reference set. If the aggregated distance is less than a threshold $\sigma$, then the triple is predicted positive and is predicted negative otherwise. The threshold $\sigma$ is tuned on the validation data. Moreover, in our experiments, the best results were obtained by the \textsc{Min} strategy and hence reported.
The results of triple classification are shown in Table \ref{tab:triple-classification}.  The results for all methods have been taken from the cited papers in the Table, except for 
TransE, and DistMult whose results are taken from the papers \cite{Yankai2015-transR} and \cite{Yao2019-kgbert}, respectively. Our approach, GilBERT, has the best performance on average and on the FB13 dataset and is second best after KG-BERT for the WN11 dataset.

\begin{table}[!ht]%\setlength\belowcaptionskip{-25pt}
	\centering
	\setlength{\tabcolsep}{6pt}

\begin{tabular}{l|ll|l}
\hline
{Method} & WN11 & FB13          & Avg. \\ \hline
%NTN                                                & 70.6 & 87.2 &78.9     %         \\
TransE    \cite{Bordes2013-transee}                                            & 75.9 & 81.5 & 79.2              \\
TransH \cite{Zhen2014-transH}                                     & 78.8 & 83.3   & 87.7            \\
TransR \cite{Yankai2015-transR}                                               & 85.9 & 82.5 & 83.9          \\
DistMult \cite{yang2015-DistMult}                                             & 87.1 & 86.2   &86.7             \\
%ConvKB \cite{Nguyen-convkb-2018}                                               & 87.6 & 88.8 & 88.2               \\
DOLORES + ConvKB \cite{wang-2020-dolores}                                             & 87.5 & 89.3   &88.4           \\
KG-BERT  \cite{Yao2019-kgbert}                                             & \textbf{93.5} & 90.4    &91.9               \\
R-MeN \cite{Nguyen-2020-RMeN}                                            & 90.5 & 88.9  & 89.7                 \\
GilBERT                                         & 92.7  & \textbf{92.5} &\textbf{92.6}      \\
\hline
\end{tabular}
\caption {Triples classification Accuracy results (in \%) reported on the WN11, FB13 test sets.}
\label{tab:triple-classification}
\end{table}

%We used a batch size of 16, Adam optimizer with learning rate $2e−5$, and a linear learning rate warm-up over 10% of the training data. Our default strategy of aggregation for the evaluation is \textsc{MEAN}. 

\noindent\textbf{Relation Prediction}
This task is to predict the relation that holds between two given entities, i.e., (h, ?, t). 
For this task, the clusters are created based on relations. The network is optimized to give small distances for embeddings of partial facts that belong to the same relation and large distances for relationally unrelated ones. We create five positive and five negative samples per anchor in the experiments.
% give an example of closest properties for a property in the FB15K dataset.

%Hence, we can used the trained network to assess the similarity of a given test data with the embedding of other partial facts whose relations are known (training points). 
\begin{table}[!ht]%\setlength\belowcaptionskip{-25pt}
	\centering
	\setlength{\tabcolsep}{6pt}

	\label{tab:relation-pred}
\begin{tabular}{l|ll}
\hline
{Method} & MR  & Hits@1        \\ \hline
TransE  \cite{Bordes2013-transee}                                              & 2.5 & 84.3          \\
TransR   \cite{Yankai2015-transR}                                             & 2.1 & 91.6          \\
ProjE (listwise) \cite{shi-2017-projE}                                              &  \text{1.2}   &   95.7            \\
TKRL  \cite{Xie-2016-tkrl}          & 1.7 & 92.8
\\
DKRL(CNN) \cite{Xie_2016_DKRL}                                                 & 2.5 & 89.0 
\\
DKRL (CNN) + TransE \cite{Xie_2016_DKRL}                                    &   2.0  & 90.8               \\
KG-BERT \cite{Yao2019-kgbert}                                              &   \text{1.2}  & \textbf{96.0}              \\
GilBERT                                          &   \text{1.3}  & 92.0\\
\hline
\end{tabular}
\caption{Relation prediction MR and HITs@1 results on FB15K dataset.}
\end{table}

For the evaluation, the correct relation $r$ is determined given its head and tail. For this purpose, the $PF_{ht}$'s embedding $E_{test}$ is derived. The $k$ closest points in the embedding space (using all training points in the reference set $\Delta$) to $E_{test}$ are computed. The true relations of the closest points are known from the training data. We propose two different strategies for assigning a relation to our test point of interest. The strategy \textsc{Min} assigns the relation of the closest point to $E_{test}$. And the strategy \textsc{K-mode} chooses the relation that has most frequently appeared among the k-closest points of $E_{test}$.

The results of the relation prediction task on the FB15K dataset are reported in Table 3. All results are taken from their respective paper, except TransE and TransR whose results are taken from \cite{Xie-2016-tkrl}. The evaluation metrics used are Hits@1 (proportion of correct relations in top 1 ranking) and Mean rank $MR$. 
Note that the reported results are after filtering. If a suggested triplet is already known to be true, i.e., if it already exists in the train, validation, or test data of the knowledge graph, it is not correct to rank it among the suggestions.
The best results with GilBERT, tuned on the validation data, were obtained using the \textsc{K-mode} strategy with K being set to 10. The Hits@1 results obtained by our approach are comparable to the state-of-the-art, without integrating any external information, with a very competitive Mean Rank.

Moreover, as the introduction states, GilBERT is more adapted to a few-shot learning setting than KG-BERT, which tackles relation prediction as a classification task. We have investigated this by comparing the Hits@1 results of the two approaches on long-tail relation, trying relations that have less than 10, 15, 20, 25, and 30 facts belonging to them in the training data and hence 
covering 37.8, 43.2, 48.4, 51.8, and 54.9 percent of the relations in the training data respectively. The Hits@1 results on these proportions of relations in the test data are plotted in Figure 2 for KG-BERT and GilBERT. This plot shows that GilBERT is more robust at relation prediction for relations that have been seen only a few times during training.

\begin{figure}
\centering
\label{fig:few-shot}
\begin{tikzpicture}
\begin{axis}[
    xlabel={Percentage of few-shot relations on the training data},
    ylabel={Filtered Hits@1 results on test data},
    xmin=35, xmax=72,
    ymin=20, ymax=80,
    xtick={35, 40, 45, 50, 55, 60, 65, 70, 75},
    ytick={0, 20, 40, 60, 80,100},
    legend pos=south east,
    ymajorgrids=true,
    grid style=dashed,
]

\addplot[
    color=blue,
    mark=square,
    ]
    coordinates {
    (37.8, 49.7)(43.2, 55.8)(48.4,61.4)(51.8,62.3)(54.9,64.0)(70.1, 73.2)%(87.4,74.8)
    };

\addplot[
    color=red,
    mark=star,
    ]
    coordinates {
    (37.8, 28.9)(43.2, 44.1)(48.4,53.2)(51.8,55.3)(54.9,62.0)(70.1, 75.5)%(87.4, 86.1)
    };
    \legend{GilBERT, KG-BERT}   
\end{axis}
\end{tikzpicture}
\caption{Comparison of KG-BERT and GilBERT hits@1 results on long-tail relations of the training data.}
\end{figure}
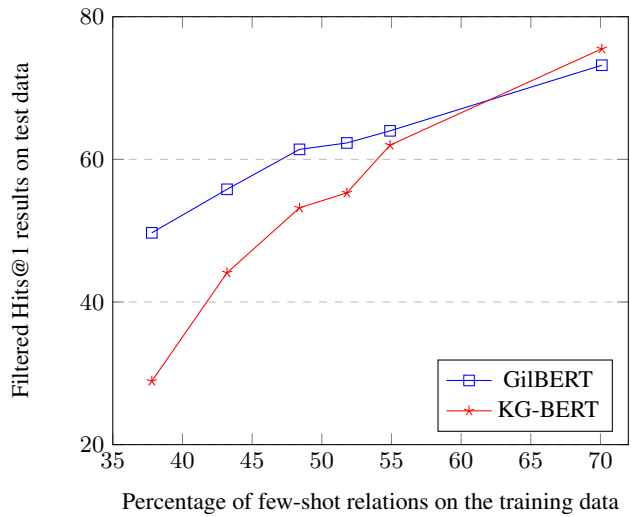

To elaborate more, we look at 664 relations (almost half of the relations) with less than 20 facts, comprising 8.4\% of FB15K instances. Our Hits@1 result on the test data involving these properties is 64\% (best results obtained with \textsc{Min} strategy) compared to KG-BERT, which gives a Hits@1 of 53\%.

%For the CoDEx dataset, the results of the other approaches were not available to compare, yet we have achieved a Hits@1 result of 92.0\% on this dataset. While having run KG-BERT with its default parameters on this task has yielded a Hits@1 of 88.8\%.  

\section{Conclusion}
\label{sect::conclusion}

This paper proposed GilBERT, a new approach that fine-tunes a triplet network of pre-trained transformer-based language models by passing textual sequences created from facts. We explained how our evaluation paradigm which relies on a very efficient spatial semantic search technique has less overhead than previous approaches and is more suited for few-shot learning settings. Finally, we reported our results on two refinement tasks.
In future work, we aim to try our approach for other KG refinement tasks such as link prediction, evaluating more benchmark datasets, and observing the impact of adding external information to training.

%,can reduce the overhead of calculating the likelihood of all unobserved facts through substitution and be more adapted to few-shot learning settings. 
%Finally, we reported the results of our approach on two refinement tasks on 3 benchmark datasets.

\vspace{0.3cm}
\noindent {\bf Acknowledgements}: 
This work has been supported by the project PSPC AIDA: 2019-PSPC-09 funded by BPI-France.

\bibliographystyle{splncs04}
\bibliography{src/references}

\end{document}